\documentclass[authoryear]{llncs}
\usepackage{graphicx}
\usepackage{amsmath,amssymb,amsfonts}
\usepackage{algorithm2e}
\usepackage{algorithmic}
\usepackage{longtable}
\usepackage{makecell}
\usepackage{subcaption}
\usepackage{booktabs}
\begin{document}
\title{Consensus Based Multi-Layer Perceptrons for
Edge Computing}

\author{Haimonti Dutta\inst{1} \and
Nitin Nataraj\inst{2}\thanks{This work was done when the author was a student at the State University of New York at Buffalo.} \and
Saurabh Amarnath Mahindre\inst{3}}
\authorrunning{Dutta et al.}
%
\institute{Department of Management Science and Systems,\\ 
The State University of New York, Buffalo, NY 14260.\\
\email{haimonti@buffalo.edu}
\and
o9 Solutions, Inc.,\\
1501 Lyndon B. Johnson Freeway, \\
Dallas, Texas 75234. \\
\email{nitin,nataraj@o9solutions.com}
 \and
Institute for Computational and Data Sciences, \\
The State University of New York, Buffalo, NY 14260.\\
\email{smahindr@buffalo.edu}}

\maketitle              
\begin{abstract}
In recent years, storing large volumes of data on distributed devices has become commonplace. Applications involving sensors, for example, capture data in different modalities including image, video, audio, GPS and others. Novel algorithms are required to learn from this rich distributed data. In this paper, we present  \emph{consensus} based multi-layer perceptrons for resource-constrained devices. Assuming nodes (devices) in the distributed system are arranged in a graph and contain vertically partitioned data, the goal is to learn a global function that minimizes the loss. Each node learns a feed-forward multi-layer perceptron and obtains a loss on data stored locally. It then gossips with a neighbor, chosen uniformly at random, and exchanges information about the loss. The updated loss is used to run a back propagation algorithm and adjust weights appropriately. This method enables nodes to learn the \emph{global} function without exchange of data in the network. 
Empirical results reveal that the consensus algorithm converges to the centralized model and has performance comparable to centralized multi-layer perceptrons and tree-based algorithms including random forests and gradient boosted decision trees.

\keywords{multi-layer perceptron  \and gossip \and consensus \and distributed learning.}
\end{abstract}
\section{Introduction}
Emerging architectures designed to store and analyze large volumes of data make use of large scale, distributed processing paradigms \cite{Bekkerman_11a}. These include the mega-scale cloud data-centers and resource constrained devices, such as the Internet of Things (IoT) and mobile devices. While the cloud can be used for executing large scale machine learning algorithms on large volumes of data, such algorithms exert severe demands in terms of energy, memory and computing resources, limiting their adoption for resource constrained, network edge devices. The new breed of intelligent devices and high-stake applications (drones, augmented/virtual reality, autonomous systems, etc.), require distributed, \emph{low-latency} and \emph{reliable} machine learning at the wireless network edge. Thus computing services have now started to move from the cloud to the edge. 

 Deep learning-based intelligent services \cite{Wang_2020} and applications have become prevalent. However, their use in edge computing devices has been somewhat limited due to the following  reasons: (a) Cost: Training and inference of deep learning models in the distributed infrastructures requires consumption of large amount of network bandwidth. (b) Latency: The access to data and services is generally not guaranteed and delay is not short enough for time-critical applications.
(c) Reliability: Most distributed computing applications rely on wireless communications and backbone networks for connecting users to services, but intelligent services must be highly reliable, even when network connections are lost
(d) Privacy: The data required for deep learning may involve private information, and privacy protocols need to be adhered. The current state of distributed deep learning systems on edge devices leaves much to be desired. 

In this paper, we address this shortcoming by developing multi-layer perceptrons for \emph{resource constrained} edge devices. When compute power is abundant and devices are not resource-constrained, deep neural networks can be trained using the DistBelief framework \cite{Dean_12} with model parallelism within (via multi-threading) and across machines (via message passing). 
Aside from the fact that a parallel architecture has a single point of failure and therefore often unsuitable for adoption in resource constrained leaderless environments, synchronization requirements lends these algorithms even more unsuitable for use on edge devices. 
Our algorithm operates on \emph{peer-to-peer computing} environments and as such interweaves local learning and label propagation \cite{Zhou_03a}. Specifically, it optimizes a trade-off between smoothness of the model parameters over the network on the one hand and the model's local learning on the other.
It has similarities to collaborative learning of personalized (peer-to-peer) models over networks \cite{Bellet_18a} -- however, unlike them, the work presented here learns the global function in the network, instead of solitary, local models. 

Finally, it must be pointed out that this work explicitly considers \emph{vertically} partitioned data or the setting in which features are distributed across nodes. Recent work on large scale distributed deep networks has primarily focused on horizontal partitions (for e.g. cross-data silo Federated Learning \cite{Kairouz_19a}) where-in all features are observed at the nodes \cite{Blot_19a,McMahan_17a} and a centralized parameter server updates models. Our work is closely related to the cross-silo Federated Learning \cite{Yang_19a,Kairouz_19a} model, except that the single point of failure parameter server in those models  is replaced with a peer-to-peer architecture. This seemingly minor change has far reaching implications -- it removes the need for synchronization with the parameter server at every iteration of the algorithm. 

This paper is organized as follows: 
Section~\ref{use} describes use cases for consensus based multi-layer perceptrons; Section~\ref{related} describes related work; Section~\ref{alg} provides details of the algorithm and empirical results are presented in Section~\ref{emp}. Section~\ref{conc} concludes the paper. 


\section{Use Cases For Learning Consensus Based Multi-layer Perceptrons}
\label{use}
We motivate the need to develop consensus based multi-layer perceptrons by describing the following applications: 
\begin{itemize}
\item \textbf{Medical Diagnosis:} Collaborations amongst health entities\footnote{https://featurecloud.eu/about/our-vision/} on mobile devices \cite{Gupta_18a} require examination of different modalities of patient data such as Electronic Health Records (EHR), imaging, pathology results, and genetic markers of a disease. 
\item \textbf{Drug Discovery: }The pharmaceutical industry requires platforms that enable drug discovery using private and competitive Drug Discovery related data and hundreds of TBs of image data\footnote{\url{https://cordis.europa.eu/project/id/831472}}.
\item \textbf{Autonomous Vehicles: }Google, Uber, Tesla, and many automotive companies have developed autonomous driving systems. 
Applications (such as forward collision warning, blind spot, lane change warnings, and adaptive cruise control.) are time critical and require real time learning and updates from individual vehicles \cite{Provodin_16}. 
\item \textbf{Home Sensing:} In home monitoring and sensing applications \cite{Huerta_16a} non-intrusive load monitoring systems are used to study fluctuations in signals. 

\item \textbf{Manufacturing Operations:} which requires industrial data that is inter-operable and scalable\footnote{\url{https://musketeer.eu/project/}}.
In applications of this genre, the sensors and IoT devices collect data at different time points often from different locations and these are then subjected to analysis. 
\end{itemize}

\section{Related Work}
\label{related}
Scalable algorithms for deep learning have been explored in several papers in recent years. We discuss related work which make use of two different architectures: (a) Parallel -- which ensures the presence of a master to control slave workers and (b) Distributed which is a fully decentralized, peer-to-peer architecture without the need for a master. 

\noindent \textbf{Parallel DNN Algorithms: }A large proportion of the research in this domain has focused on data parallelism and the ability to exploit compute power of multiple slave workers, with a single master controlling the execution of slaves. McDonald et al. \cite{McDonald_10a} present two different strategies for parallel training of structured perceptrons and use them for named entity recognition and dependency parsing. TernGrad \cite{Wen_17a} uses ternary levels $\{-1, 0, +1\}$ to reduce overhead of gradient synchronization and communication. 
DoReFa-Net \cite{Zhou_16a} train convolutional neural networks that have low bit width weights, activations and gradients. 
Seide et al. \cite{Seide_14a} show that it is possible to quantize gradients aggressively during training of deep neural networks using SGD making it feasible to use in data parallel fast processors such as GPUs. Quantized SGD (QSGD) \cite{Alistarh_17a} explores the trade-off between accuracy and gradient precision. 
A slightly different line of work \cite{Zhang_16a} explores the utility of asynchronous Stochastic Gradient Descent algorithms suggesting that if the learning rate is modulated according to the gradient staleness, better theoretical guarantees for convergence can be established than the synchronous counterpart. 


\noindent \textbf{Distributed DNN Algorithms:} In the fully decentralized setting, \cite{Jiang_17a} present a consensus-based distributed SGD (CDSGD) algorithm for collaborative deep learning over fixed topology networks that enables data parallelization as well as decentralized computation. 
Sutton et al. \cite{Sutton_09a} explore neural network architectures in which the structures of the models are partitioned prior to training. 
Partitioning of deep neural networks have also been studied in the context of distributed computing hierarchies such as the cloud, end and edge devices \cite{Teera_17a}. 
Gupta et al. \cite{Gupta_18a} present an algorithm for training DNNs over multiple data sources. 
The research described above fundamentally differ from the material presented in this paper in that our consensus algorithm relies on both model and data partitioning to construct local multi-layer perceptron models which can independently learn global information. 


%

\section{Distributed Multi-layer Perceptrons}
\label{alg}
We present the consensus based multi-layer perceptron algorithm in this section. 

In the distributed setting, let $M$ denote an $N \times n$ matrix with real-valued entries.  
This matrix represents a dataset of $N$ tuples of the form $x_i \in \mathbb{R}^n, 1 \le i \le N$. Each tuple has an associated label $y_i = \{+1, -1\}$.
Assume this dataset has been \emph{vertically}\footnote{This implies that all the nodes have access to all $N$ tuples but have limited number of features i.e. $n_i \le n$.} distributed over $m$ nodes $S_1, S_2, \cdots, S_m$ such that node $S_i$ has a data set $M_{i} \subset M, M_{i}: N \times n_i$ and each $x_j \in M_{i}$ is in $\mathbb{R}^{n_i}, n_i \le n$. Thus, $M = M_1 \cup M_2 \cup \cdots \cup M_m$ denotes the concatenation of the local data sets. The labels are shared across all the nodes. The goal is to learn a deep neural network on the global data set $M$, by learning local models\footnote{We assume that the models have the same structure i.e. the same number of input, hidden and output layers and connections.} at the nodes, allowing exchange of information among them using a gossip based protocol \cite{Kempe_03a,Demers_87a} and updating the local models with new information obtained from neighbors. This ensures that there is no actual data transfer amongst nodes. 

\noindent \textbf{Model of Distributed Computation. } The distributed algorithm evolves over discrete time with respect to a ``global" clock\footnote{Existence of this clock is of interest only for theoretical analysis}. Each node has access to a local clock or no clock at all. Furthermore, each node has its own memory and can perform local computation (such as estimating the local weight vector). It stores $f_i$, which is the estimated local function. Besides its own computation, nodes may receive messages from their neighbors which will help in evaluation of the next estimate for the local function. 

\noindent \textbf{Communication Protocols. }Nodes $S_i$ are connected to one another via an underlying communication framework represented by a graph $G (V, E)$, such that each node  $S_i \in \{S_1, S_2, \cdots , S_m\}$ is a vertex and an edge $e_{ij} \in E$ connects nodes $S_i$ and $S_j$. Communication delays on the edges are assumed to be zero. 

\noindent  \textbf{Distributed MultiLayer Perceptron (DMLP): } Assume that each node $S_t$ has a simple model of a fully connected multi-layer perceptron with sigmoid activations for hidden layers and sigmoid for the output layer. The network is called $\mathcal{N}_t$. It has $L$ layers -- the $0^{th}$ is the input layer, followed by $(L-1)$ hidden layers and the $L^{th}$ layer is the output layer. Let $r_i$ denote the number of units in the $i^{th}$ layer (note that $r_0 = n_i$ and $r_L = 1$). 


\noindent \textbf{Feed-forward Learning: } Let  $\omega_{ij}^k$ denote the weight from $i^{th}$ node of $(k-1)^{th}$ layer to $j^{th}$ node of $k^{th}$ layer, $a_{j}^k$ is the weighted sum of inputs from the previous layer to the $j^{th}$ node of $k^{th}$ layer, $o_{j}^k$ is the output of $j^{th}$ node of $k^{th}$ layer, $b_{j}^k$ is the bias to $j^{th}$ node of $k^{th}$ layer. The feed-forward step for the first node of the first hidden layer can then be written as: $a_{1}^1=b_{1}^1 + x_{1}^0 * \omega_{11}^1 + x_{2}^0 * \omega_{21}^1 + ... + x_{n_i1}^0 * \omega_{n_i1}^1$. The output $a_{1}^1$ is given by: $o_{1}^1 = \sigma(a_{1}^1)$. So, the output of $j^{th}$ node of $k^{th}$ layer is, $o_{j}^k = \sigma(a_{j}^k)$ where, $a_{j}^k = b_{j}^k + (\sum_{i=1}^{r_{k-1}} o_{i}^{k-1}*w_{ij}^k)$.
The output from the network $\mathcal{N}_t$ is given by
$	\hat{y}^{\mathcal{N}_t}_i = \sigma(a_{1}^L)$.
The local loss at node $S_i$ is then given by $ \mathcal{L}_t= \frac{1}{2} \sum_{i=1}^{N} (y_i - \hat{y}^{\mathcal{N}_t}_i)^2$, assuming squared loss.

 \noindent \textbf{Gossip: }Node $S_t$ selects uniformly at random, a neighbor $S_u$ with whom it wishes to gossip. Both $S_t$ and $S_u$ have computed their local losses. When gossiping each node updates its current local loss with  $\mathcal{L}_{gossip} = \frac{\mathcal{L}_t + \mathcal{L}_u}{2}$. This new loss is used for back propagation at both nodes $S_t$ and $S_u$.
 
\noindent \textbf{Back propagation: }The back propagation algorithm learns the weights for a multi-layer network, given a network with a fixed set of units and interconnections. It employs gradient
descent to attempt to minimize the squared error between the network output values and the target values for these outputs. We use the new loss ($\mathcal{L}_{gossip}$) obtained after gossiping with a neighbor, in-place of the local loss ($\mathcal{L}_t$), for our back propagation phase. This modification helps the local node $S_t$ to incorporate information about the loss from its neighbor $S_u$ into its back propagation learning phase, thereby helping to minimize the \emph{global} loss instead of the local loss. This is a crucial step in our algorithm.  The local loss at node $S_t$ after gossip is then given by $ \mathcal{L}_{gossip}= \frac{1}{2} \sum_{i=1}^{N} (y_i - \frac{\hat{y}^{\mathcal{N}_t}_i + \hat{y}^{\mathcal{N}_u}_i}{2})^2 = \frac{1}{2} (\mathbf{y - y_{gossip} })^2; \mathbf{y_{gossip}}=  \frac{\hat{y}^{\mathcal{N}_t}_i + \hat{y}^{\mathcal{N}_u}_i}{2}$ where the bold fonts are used to represent the loss vectors. Algorithm~\ref{alg:DMLP} presents the steps of the DMLP algorithm.

\begin{algorithm}[!h]
\small
{
\SetKwData{Left}{left}\SetKwData{This}{this}\SetKwData{Up}{up}
\SetKwFunction{Union}{Union}\SetKwFunction{FindCompress}{FindCompress}
\SetKwInOut{Input}{input}\SetKwInOut{Output}{output}

\KwIn{$N \times n_i$ matrix at each node $S_i$, $G(V, E)$ which encapsulates the underlying communication framework, $T: $ no of iterations }
\KwOut{Each node $S_i$ has a multilayer perceptron network $\mathcal{N}_i$ }
\BlankLine

 %
\For{t = 1 to T}
 {
  (a) Node $S_i$ uses the network $\mathcal{N}_i$ for feedforward learning and locally estimates the loss on N instances\; 
  (b) Node $S_i$ gossips with its neighbors $S_j$ and obtains the loss from the neighbor\; 
  (c) \textbf{Gossip: }node $S_i$ averages the loss between $S_i$ and $S_j$ and sets this as the new loss\; 
  (d) Perform backpropagation on the current node and the neighbor node using the gossiped loss; Update the weight vectors in each layer using Stochastic Gradient Descent (SGD)\; 
  (e) If there is no significant change in the local weight vectors, STOP \; 
 } 
\caption{Distributed Multilayer Perceptron Learning (DMLP)}
\label{alg:DMLP}
}
\end{algorithm}

\noindent \textbf{Discussion: }Some interesting aspects of our algorithm are: (a) The algorithms presented in the above section are called anytime algorithms \cite{SZ_93}. Anytime algorithms are those whose quality of results change gradually as computation time increases. 
At a given time a node may be interrupted to obtain an estimate of the performance. 
(b) Algorithm~\ref{alg:DMLP} can be  extended to other kinds of loss functions (such as cross-entropy, softmax) and activations (such as linear, tanh). 
(c) The number of hidden layers of the multi-layer perceptron can be incremented as required by a node, without the need for any algorithmic changes. 

\section{Empirical Results}
\label{emp}
The empirical results demonstrate the utility of the DMLP algorithm. We examine the following questions: (a) Is there empirical support for the conjecture that the performance of the distributed model is better than that of the centralized model? (b) Does the distributed model empirically converge to the centralized one? 
(c) How does the performance of the proposed method compare to feature subspace learning methods such as Random Forests \cite{Breiman_01a} and tree boosting algorithms (such as XGBoost \cite{Chen_16a})?
The answers to the above questions are explored using the data sets\footnote{For MNIST, a balanced binary classification data set was produced using digits 0 and 9; For CIFAR, to convert to a binary classification problem, we assign classes 0-4 the label 0, and classes 5-9, label 1.} shown in Table~\ref{tab:dataset} \cite{Guyon_04a}. 
\begin{table}[t]
\begin{center}

 \begin{tabular}{|c | c | c | c|} 
 \hline
 \textbf{Dataset} & \textbf{No. Train} & \textbf{No. Test} & \textbf{No. Features}\\ 
 \hline
 Arcene & 100 & 100 & 10000 \\ 
 \hline
 Dexter & 300 & 300 & 20000 \\ 
 \hline
 Dorothea Bal. & 156 & 68 & 100000 \\
 \hline
 Gisette & 6000 & 1000 & 5000 \\ 
 \hline
 Madelon & 2000 & 600 & 500\\
 \hline
 MNIST Bal. & 11702 & 1948 & 784\\ 
 \hline
 HT Sensor & 14560 & 3640 & 10 \\ 
 \hline
 CIFAR-10 & 50000 & 10000 & 3072 \\ 
 \hline
\end{tabular}
\end{center}
\caption{\label{tab:dataset} Characteristics of the datasets used for empirical analysis.}
\end{table}

The experimental process is as follows:(a) The Peersim simulator \cite{peersim} is used to construct a fully connected graph of $10$ nodes. Each node can independently store vertically partitioned data. (b) The total number of features in the train data is split into $10$ roughly equal parts. Each node is assigned the data with the corresponding split containing all the examples but only those features it has been assigned. (c) Each node builds a local neural network model. The local loss vector is generated. (d) Each node selects a neighbor uniformly at random according to the underlying distributed graph, and exchanges the local loss vector with its neighbor. The new loss vector is computed as the average of its own loss vector and that of the neighbor's. (e) Each node participates in back propagation using the new loss generated after gossiping with a neighbor. (f) The above process is repeated for several iterations until the nodes converge to a solution. 

\noindent \textbf{Testing the model: }Each node is provided with the test set having only those features that the node used to construct the local model. Therefore, each node can test its own performance. For experiments presented here, we construct the following hypothetical scenario: for each test sample, an average predicted probability is obtained across all nodes, and the distributed test AUC is then estimated. This is not a requirement of the algorithm, but it enables us to compare performance against benchmarks.

\begin{table*}[!h]
\begin{center}
\resizebox{\textwidth}{!}{%
{ \begin{tabular}{|l | l | l | l | l | l | l | l | l | l |}  \hline
 Dataset & \thead{No. \\ Hidden \\ Neurons (C)} & \thead{No.\\ Hidden \\ Neurons (D)} & \thead{Learning \\ Rate} & \thead{Centralized \\ AUC} & \thead{Distributed \\ AUC} & \thead{$95\%$ C. I.} & \thead{Cent. \\ Itr. ($I_C$)} & \thead{Dist. \\ Itr. ($I_D$)} \\  \hline
 Arcene & $50$ & $5$ & $0.01$ & $0.93 \pm 0.006$ & $0.92 \pm 0.01$ & $[0.88,0.96 ]$& $299 \pm 16.33$ & $365.67 \pm 233.28$  \\ \hline
Dexter & $20$ & $2$ & $0.5$ & $0.64 \pm 0.03$ & $0.86 \pm 0.02$ & $[0.83,0.89]$ &  $29 \pm 0$ & $29 \pm 0$ \\ \hline
Dorothea Bal. & $100$ & $10$ & $0.5$ & $ 0.91 \pm 0.007$ & $0.92 \pm 0.01$ & $[0.87,0.97]$ & $42.33 \pm 4.71$ & $55.67 \pm 4.71$ \\ \hline
Gisette & $200$ & $20$ & $0.05$ & $0.99 \pm 0.001$ & $0.99 \pm 0.006$ & $[0.991,0.997]$ & $155.67 \pm 9.43$ & $182.33 \pm4.71$ \\ \hline
Madelon & $50$ & $5$ & $0.1$ & $0.62 \pm 0.003$ & $0.63 \pm 0.009$ & $[0.6,0.66]$ & $462.33 \pm 73.18$ & $292.33 \pm 49.22$ \\\hline
MNIST Bal. & $100$ & $10$ & $0.5$ & $0.98 \pm 0.009$ & $0.99 \pm 0.0005$ & $[0.987,0.993]$ & $209 \pm 14.14$ & $189 \pm 16.33$ \\ \hline
HT Sensor & $20$ & $2$ & $0.1$ & $0.99 \pm 0.002$ & $0.99 \pm 0.01$ & $[0.98,0.99]$ & $149 \pm 65.32$ & $205.67 \pm 49.89$ \\ \hline
CIFAR-10 & $400$ & $40$ & $0.1$ & $ 0.72 \pm 0.001 $ & $0.71 \pm 0.001$ & $[0.71, 0.71]$  & $145 \pm 75$ & $52 \pm 9$  \\ \hline
\end{tabular}
}
}
\end{center}
\caption{\label{tab:results} Performance of the centralized (C) and distributed algorithms (D). The consensus multi-layer perceptron uses cross-entropy loss function, ReLU activation for the hidden layer, and sigmoid activation for the output layer. The results are averaged over three trials.}
\end{table*}
We measure the performance of the model by the area under the Receiver Operating Characteristic (ROC) curve \cite{rocauc} denoted by $\theta$. 
\begin{table*}[!h]
\begin{center}
\begin{tabular}{ccccc} \hline
Dataset & RF$\_$AUC & XGBoost$\_$AUC & Dist$\_$AUC & Cent$\_$AUC \\ \hline
Arcene & $0.79$ & $0.84$ & $0.92$ & $0.93$ \\
Dexter & $0.93$ & $0.95$ & $0.86$ & $0.64$ \\
Dorothea Bal. & $0.88$ & $0.89$ & $0.92$ & $0.91$ \\ 
Gisette & $0.99$ & $0.99$ & $0.99$ & $0.99$ \\
Madelon & $0.77$ & $0.71$ & $0.63$ & $0.62$ \\
MNIST Bal.& $0.99$ & $0.99$ & $0.99$ & $0.98$ \\
HT & $1$ & $0.99$ & $0.99$ & $0.99$ \\ 
CIFAR-10 & $0.67 $ & $0.71 $ & $0.71$ & $0.72$ \\ \hline
\end{tabular}
\end{center}
\caption{Comparison of the performance of the consensus algorithm to tree based algorithms Random Forest (RF) and XGBoost.}
\label{tab:RF}
\end{table*}%
The centralized algorithm is executed by assuming that the entire dataset is available at a node. In the distributed setting, a neural network is employed at each node, and is fed partial data, partitioned in the feature space. The number of total hidden neurons is kept the same for both the centralized and distributed experiments. This implies that each distributed node has roughly $\frac{\text{No. of hidden neurons in cent. model}}{\text{No. of nodes}}$ hidden neurons in its model. We tune the model(s) in each experiment by selecting different parameters for learning rate, number of hidden neurons, number of hidden layers and activation functions. 

\begin{figure*}[!t]
\begin{minipage}[b]{1.0\linewidth}
  \begin{center}
\includegraphics[width=0.98\textwidth]{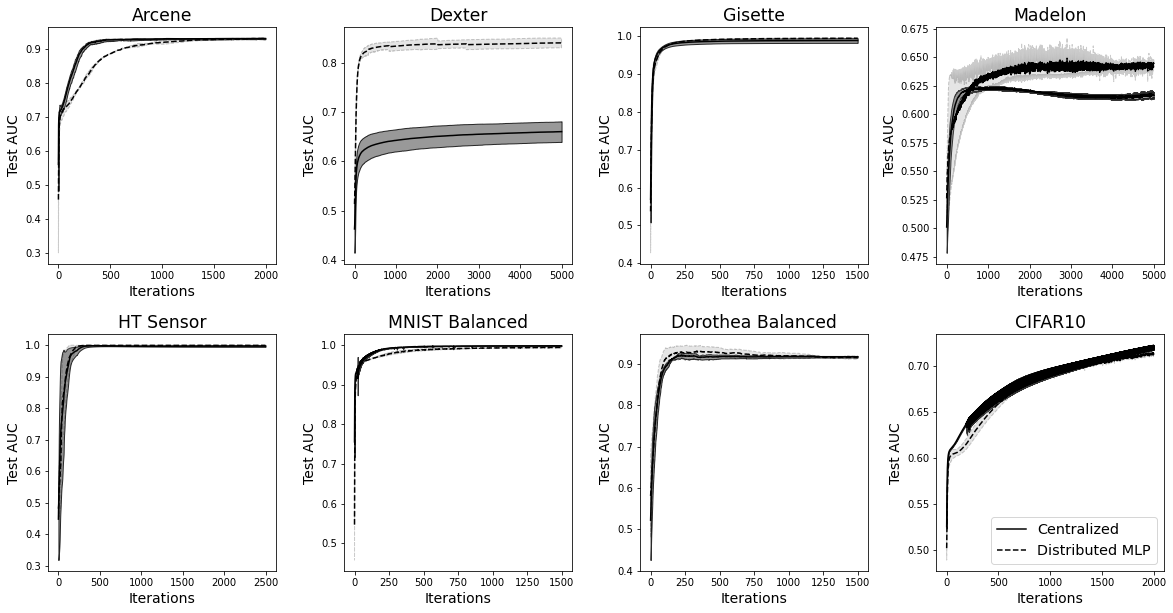}
  \end{center}
  \caption{AUC on the test sets for both centralized and distributed settings on the eight datasets discussed above. For the distributed algorithm, test AUC results averaged over three random vertical feature splits without overlap are presented.}
\label{fig:testaucplots}
\end{minipage}
\end{figure*}

The steps outlined for the distributed algorithm above were repeated for three random feature splits and the test AUC averaged over the trials. We also compute the symmetric 95\% confidence interval for distributed test AUC ($\theta_D$) and observe centralized test AUC ($\theta_C$) in relation to this interval. Figure~\ref{fig:testaucplots} shows the AUC curves for all the datasets used in this study.
\begin{table}[!h]
\begin{center}
{ \begin{tabular}{|l  | l | l | l | l |}  \hline
 Dataset & \thead{Cent \\ AUC} & \thead{Dist. w/overlap \\ AUC} & \thead{$95\%$ C. I.}  \\  \hline
Arcene & $0.93 \pm 0.01$ & $0.92 \pm 0.00$ & $[0.88,0.96 ]$   \\ \hline
Dexter &  $0.64 \pm 0.03$ & $0.81 \pm 0.02$ & $[0.77,0.84]$  \\ \hline
Doro. Bal. &  $ 0.91 \pm 0.01$ & $0.91 \pm 0.03$ & $[0.86,0.96]$  \\ \hline
Gisette & $0.99 \pm 0.00$ & $0.99 \pm 0.00$ & $[0.99,0.99]$  \\ \hline
Madelon &  $0.61 \pm 0.00$ & $0.63 \pm 0.00$ & $[0.60,0.67]$  \\\hline
MNIST &  $0.98 \pm 0.01$ & $0.99 \pm 0.00$ & $[0.99,0.99]$  \\ \hline
HT Sensor &  $0.99 \pm 0.00$ & $0.99 \pm 0.00$ & $[0.99,0.99]$  \\ \hline
CIFAR-10 & $0.72 \pm 0.001$ & $0.71 \pm 0.01$ & $[0.71, 0.71]$   \\\hline
\end{tabular}
}
\end{center}
\caption{\label{tab:results1} Performance of the centralized(C) and distributed algorithms(D) with $20\%$ overlap of features. The consensus neural network uses cross-entropy loss function, ReLU activation for the hidden layer, and sigmoid activation for the output layer. The results are averaged over three trials.  }
\end{table}
The Standard Error ($SE$) for estimated area under the ROC curve in relation to the sample size ($n$) and $\theta_D$ can be computed as described in \cite{hanley_mcneil}:

$SE = \sqrt{\frac{\theta_D(1 - \theta_D) + (n-1)(Q_1 + Q_2-2\theta_D^2)}{n^2}}$
where
$Q_1 = \frac{\theta_D}{2-\theta_D} , Q_2 = \frac{2\theta_D^2}{1+\theta_D}$.

\begin{figure*}[!h]
\begin{minipage}[b]{1.0\linewidth}
  \begin{center}
  \includegraphics[width=0.98\textwidth]{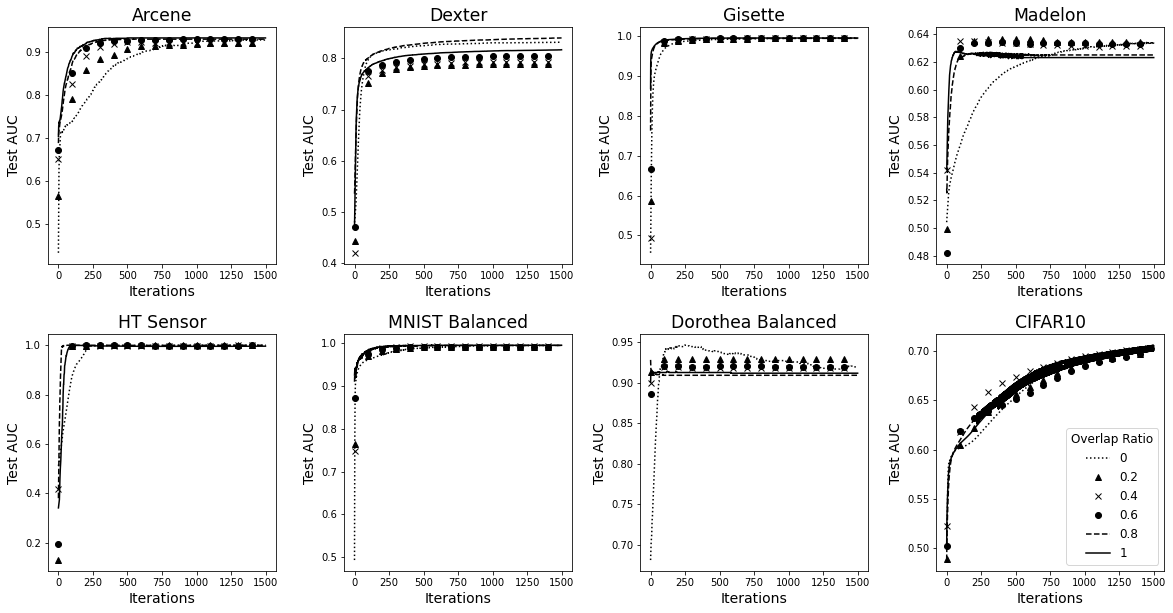}
  \end{center}
  \caption{AUC on the test sets for both centralized and distributed settings on the eight datasets discussed above with varying degree of feature overlap. For the distributed algorithm, test accuracy results averaged over three random vertical feature splits with and without overlap are presented.}
\label{fig:testOverlapplots}
\end{minipage}
\end{figure*}

Given SE, the symmetric 95\% confidence interval ($CI$) is given by $\theta_D \pm 1.96(SE)$. The centralized algorithm and distributed algorithm can be deemed approximately comparable if $\theta_C$ lies within these bounds, i.e. if $\theta_D - 1.96(SE) <= \theta_C <=\theta_D + 1.96(SE)$. In empirical studies (Table~\ref{tab:results}), it was found that the distributed algorithm obtains comparable test AUC scores to the centralized algorithm for all the datasets. 

\begin{figure*}[!h]
\begin{minipage}[b]{1.0\linewidth}
  \begin{center}
  \includegraphics[width=0.98\textwidth]{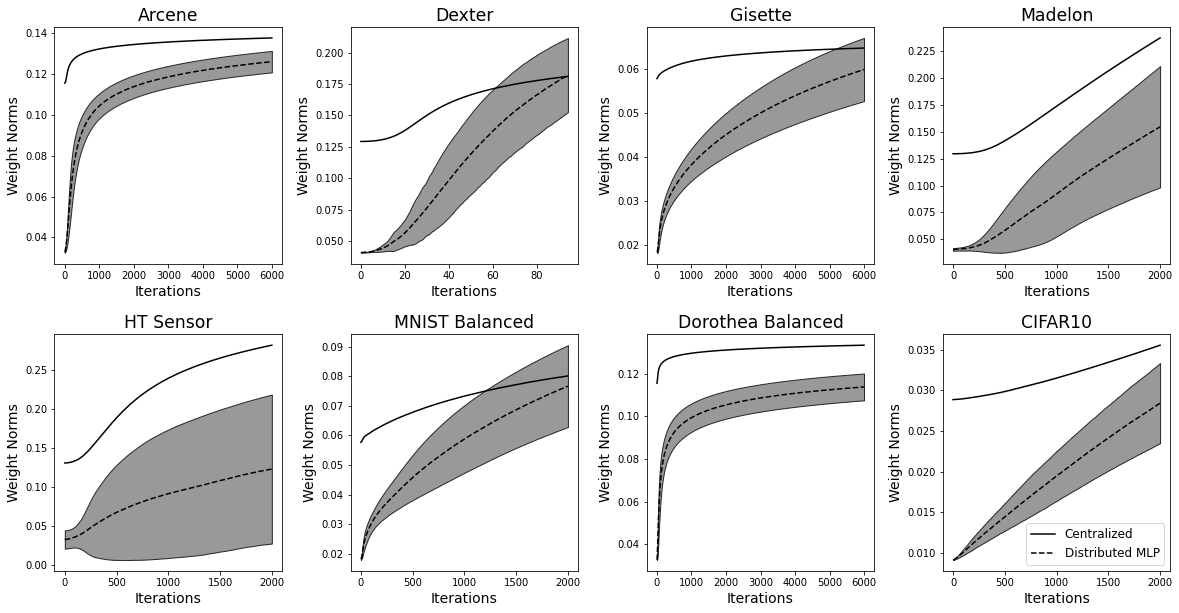}
  \end{center}
  \caption{Verification of the empirical convergence of the distributed algorithm.}
\label{fig:conv}
\end{minipage}
\end{figure*}

\noindent \textbf{Empirical Convergence of DMLP Algorithm} To test the convergence of the DMLP algorithm, we measure the difference in $L2$ norms of the normalized weight vectors in the centralized and distributed algorithms as they progress through the algorithm. Figure~\ref{fig:conv} shows that for all the data sets, this difference approaches zero, thereby supporting our conjecture that the distributed algorithm follows the behavior of the centralized one. This is a very important result, because even though the multi-layer perceptrons were trained on separate nodes with partial data, the global objective function was being minimized lending it useful for many applications on resource constrained devices.

\noindent \textbf{Effect of Overlap of Features} We study the impact of the overlap of features at each node on the performance of the consensus algorithm using the $overlap\_ratio$ parameter. An $overlap\_ratio$ of 0 indicates that the features present at one node are not present at any other node, i.e. the feature space is partitioned with mutual exclusivity. On the other hand, an $overlap\_ratio$ greater than 0, indicates that a subset of the feature space is shared among all nodes. When $overlap\_ratio$ is 1, all data is available at all nodes but model partition still exists. 
Table~\ref{tab:results1} presents the results of the experiments with $overlap\_ratio$ set to $0.2$. Figure~\ref{fig:testOverlapplots} shows the effect of variation of the $overlap\_ratio$ parameter on performance of the algorithm. In general, it is observed that when the  $overlap\_ratio$ is incremented by a factor of $0.2$, the AUC on the test set gradually improves. 
Our results reveal that in general, the overlap of features amongst nodes is beneficial and boosts the performance of the consensus algorithm. However, this behavior is not consistent for highly nonlinear datasets (such as Madelon) and those which have very large number of features (such as Dorothea) where-in the performance decreases as overlap increases and overfitting sets in. 

\noindent \textbf{Comparison with feature sub-space learning algorithms} Given that data partition at each node involves exploring a subset of the feature space, we compare the consensus algorithm to state-of-the-art tree-based algorithms which learn on feature subspaces (such as Random Forests and XGBoost). The results are presented in Table~\ref{tab:RF}. We observe that the consensus algorithm has comparable performance to RF and XGBoost in all the datasets, except Dexter and Madelon -- two particularly difficult datasets with no informative features \cite{Guyon_04a}.

\section{Conclusion and Future Work}
\label{conc}
This paper presents an algorithm for learning consensus based multi-layer perceptrons in resource constrained edge devices. The devices (nodes) are arranged in a network and contain vertically partitioned data.
Each node constructs a local model by feed forward learning, exchanges losses with a randomly chosen neighbor, averages losses and uses this new loss for back propagation. 
Empirical results on several real world datasets reveal that the consensus algorithm has performance comparable to the centralized counterpart and tree-based learning algorithms. Future work involves exploring the possibility of extending the algorithm to other kinds of neural networks and developing distributed perceptrons for multi-class classification.

\bibliographystyle{plain}
\bibliography{uai}
\end{document}